\documentclass[letterpaper, 10 pt, conference]{ieeeconf}  % Comment this line out if you need a4paper

\IEEEoverridecommandlockouts                              % This command is only needed if 
\usepackage{amsfonts}
\usepackage{amsmath}
\usepackage{graphics} % for pdf, bitmapped graphics files
\usepackage{epsfig} % for postscript graphics files
\usepackage{xcolor}
\usepackage{algorithm}
\usepackage{algpseudocode}

\title{\LARGE \bf
    Autonomous Needle Navigation in Retinal Microsurgery: \\
    Evaluation in \textit{ex vivo} Porcine Eyes
}
\author{Peiyao Zhang$^{1}$, Ji Woong Kim$^{1}$, Peter Gehlbach$^{2}$, Iulian Iordachita$^{1}$, and Marin Kobilarov$^{1}$% <-this % stops a space
\thanks{$^{1}$Peiyao Zhang, Ji Woong Kim, Iulian Iordachita, and Marin Kobilarov are with the Department of Mechanical Engineering and the Laboratory for Computational Sensing and Robotics (LCSR), Johns Hopkins University, Baltimore, MD 21211, USA 
        {\tt\small \{pzhang24, jkim447, iordachita, mkobila1\}@jhu.edu}}%
\thanks{$^{2}$Peter Gehlbach is with the Wilmer Eye Institute, Johns Hopkins University, Baltimore, MD 21211, USA 
        {\tt\small pgelbach@jhmi.edu}}%
}

\begin{document}
\maketitle
\thispagestyle{empty}
\pagestyle{empty}

%%%%%%%%%%%%%%%%%%%%%%%%%%%%%%%%%%%%%%%%%%%%%%%%%%%%%%%%%%%%%%%%%%%%%%%%%%%%%%%%
\begin{abstract}

Important challenges in retinal microsurgery include prolonged operating time, inadequate force feedback, and poor depth perception due to a constrained top-down view of the surgery. The introduction of robot-assisted technology could potentially deal with such challenges and improve the surgeon's performance. Motivated by such challenges, this work develops a strategy for autonomous needle navigation in retinal microsurgery aiming to achieve precise manipulation, reduced end-to-end surgery time, and enhanced safety. This is accomplished through real-time geometry estimation and chance-constrained Model Predictive Control (MPC) resulting in high positional accuracy while keeping scleral forces within a safe level. The robotic system is validated using both open-sky and intact (with lens and partial vitreous removal) \textit{ex vivo} porcine eyes. The experimental results demonstrate that the generation of safe control trajectories is robust to small motions associated with head drift. The mean navigation time and scleral force for MPC navigation experiments are 7.208 s and 11.97 mN, which can be considered efficient and well within acceptable safe limits. The resulting mean errors along lateral directions of the retina are below 0.06 mm, which is below the typical hand tremor amplitude in retinal microsurgery.
\end{abstract}

%%%%%%%%%%%%%%%%%%%%%%%%%%%%%%%%%%%%%%%%%%%%%%%%%%%%%%%%%%%%%%%%%%%%%%%%%%%%%%%%
\section{INTRODUCTION}

Retinal microsurgery requires precise manipulation of the delicate and non-regenerative tissue of the retina, while using high precision surgical tools. Its requirement for micrometer accuracy often exceeds the capability of even the most experienced surgeons. The success of retinal microsurgery relies on a clear view of the surgical workspace and precise hand-eye coordination while managing physiological hand tremor. Riviere et al. \cite{riviere2000study} have reported that the root mean square amplitude of hand tremor in retinal microsurgery is around 0.182 mm while the required positioning accuracy of subretinal injection can be as small as 0.025 mm \cite{zhou2018precision}. Risks that may lead to damage of the retina and sclerotomy port must be reduced to avoid irreversible complications to the eye tissue. One challenge in retinal microsurgery is prolonged surgery duration. Loriga et al. \cite{loriga2018postoperative} have shown a positive association between postoperative pain and duration of surgery. 
% Khwarg et al. \cite{khwarg1987incidence} reported 39\% and 0.9\% incidences of light retinopathy in cataract surgery with the operating time greater than and less than 100 minutes respectively. 
Experimentally induced light toxicity increase significantly after 13 minutes \cite{williamson2008introduction}. It is also thought that muscular and mental fatigue develop with prolonged operating times that exacerbate fine motor control and hand tremor concerns \cite{slack2008effect}. 
\begin{figure}[thpb]
  \centering
  \includegraphics[scale=0.08]{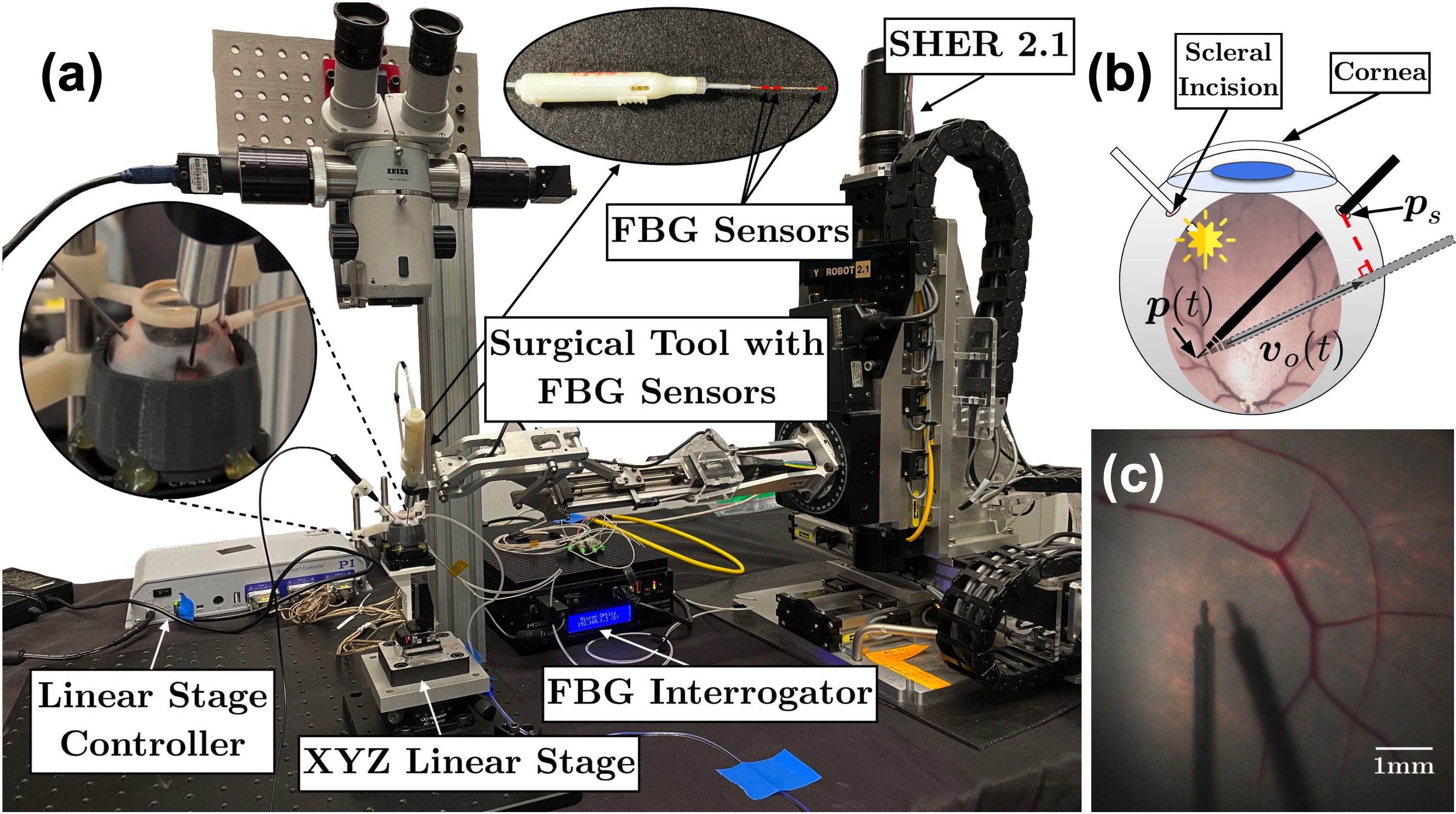}
  \vspace{-8pt}
  \caption{Robotic system overview: (a) Scleral force measurement setup. (b) Scleral incision point and cornea. The red-dashed line is set to zero as the scleral constraint. (c) The top-down view of an intact porcine eye.}
  \label{fig:FBG_setup_scleral_incision_intact_eye}
  \vspace{-18pt}
\end{figure}
Due to physical limitations of the human surgeon and demands for short surgery duration, robot-assisted technology for autonomy and safety is necessary.

Various works have focused on robotic applications for retinal microsurgery. During these studies, artificial eye phantoms are commonly used due to their fixed and simple geometry. However, the use of porcine eyes, which are used in this study, imposes more challenges compared to artificial eye phantoms, since biological tissues have more irregular and unpredictable geometry. Furthermore, the additional physical constraints applied to the Remote-Center-of-Motion (RCM) at the scleral incision point (Fig. \ref{fig:FBG_setup_scleral_incision_intact_eye}b) will make the robot system more complicated. In this work, we implement both open-sky and intact \textit{ex vivo} porcine eye experiments. We have shown that by using this experimental setup (Fig. \ref{fig:FBG_setup_scleral_incision_intact_eye}a), we can achieve high image resolution, and a field of view approaching that of a stereo microscope, Fig. \ref{fig:FBG_setup_scleral_incision_intact_eye}c. 

\begin{figure}[thpb]
  \centering
  \includegraphics[scale=0.113]{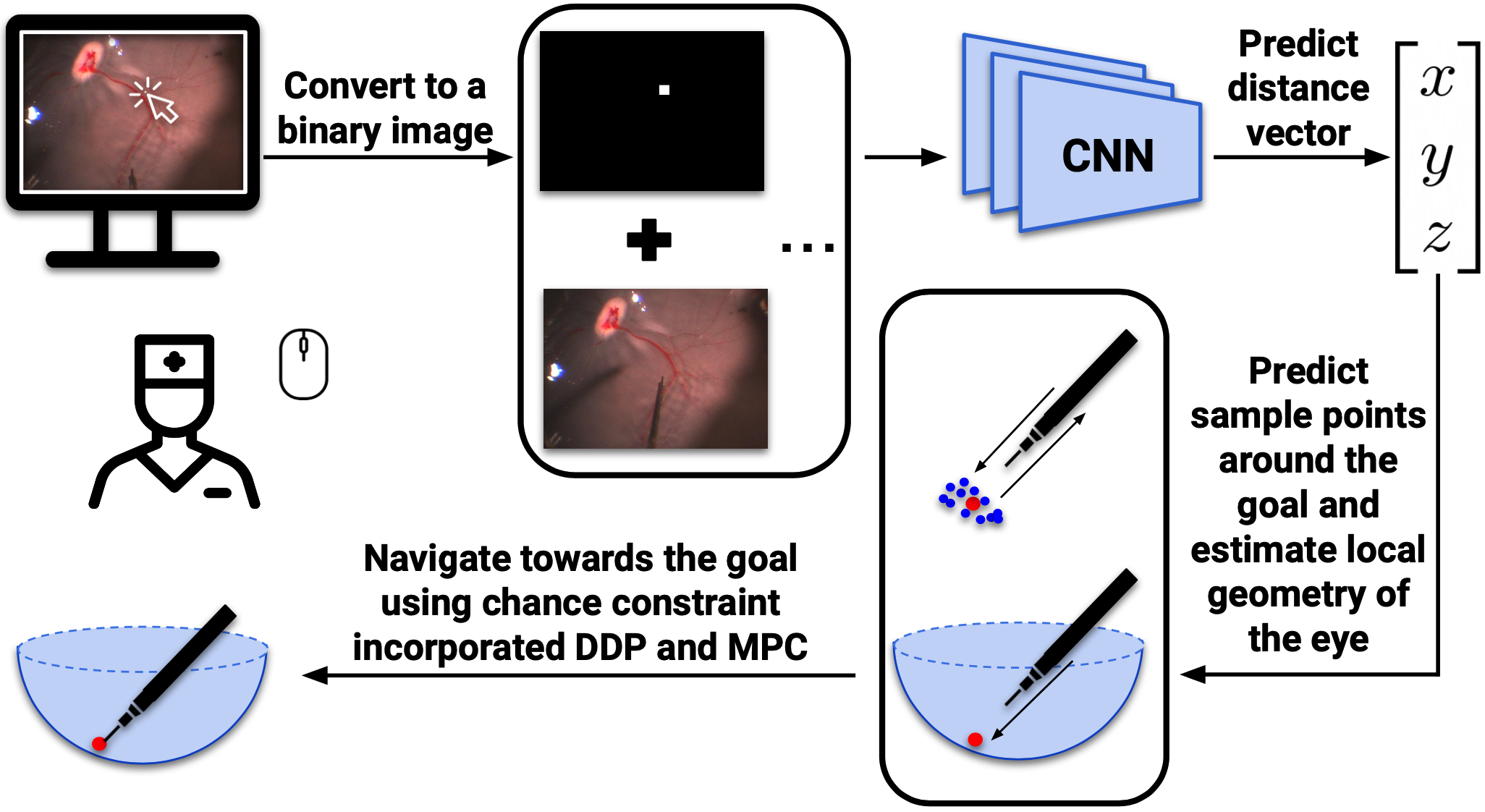}
  \vspace{-10pt}
  \caption{Workflow of autonomous navigation process using MPC.}
  \label{fig:workflow}
  \vspace{-18pt}
\end{figure}

Our previous work \cite{kim2020autonomously}, \cite{kim2021towards} demonstrated that autonomous navigation was possible using predicted depth information and optimal control. We used a deep neural network to assess depth perception, and predict the distance vector based on a top-down microscope image and a user-defined goal position, Fig. \ref{fig:workflow}.
% \begin{figure}[thpb]
%   \centering
%   \includegraphics[scale=0.15]{Figs/FBG_setup.jpg}
%   \caption{Scleral force measurement setup.}
%   \label{fig:FBG_setup}
%   \vspace{-18pt}
% \end{figure}
Furthermore, we proposed a method \cite{zhang2021towards} that combined real-time geometry estimation and chance constraint to generate a safer autonomous navigation process. However, all these works were validated using only silicone eye phantoms. Moreover, we only dealt with the simplest situation in which the eye was static.

The present work extends and improves prior results in several ways. First, we demonstrate improved (in terms of completion time, accuracy, and safety) autonomous navigation. To better estimate the local geometry of the retina, we add a regularization term to the geometry fitting equation. We improve our chance constraint formulation through linearizing all ellipsoid parameters. By the use of MPC and chance constraint incorporated Differential Dynamic Programming (DDP) \cite{mayne1966second}, the safety and navigation speed are thereby guaranteed. This could potentially alleviate the surgeon of any burden incurred by navigating the tool to the desired position, and allow the surgeon to focus on the main surgical objectives.
Second, we evaluate the robotic system in a more realistic setting using both open-sky and intact \textit{ex vivo} porcine eyes.
%The use of DDP also makes it possible to introduce obstacles in the future if necessary.
Third, we show that the system is sufficiently robust to navigate to the updated goal position during simulated small head drift. To prove the efficiency of our scleral constraint, we measure the scleral force using a Fiber Brag Grating (FBG) sensors integrated surgical tool (Fig. \ref{fig:FBG_setup_scleral_incision_intact_eye}a) developed by He et al. \cite{he2014multi}. The results show that the scleral force can be maintained at a safe level.

\section{RELATED WORK}
\subsection{Hand Tremor and Scleral Force Reduction}
Numerous studies have focused on reducing hand tremor and scleral forces. Our group from Johns Hopkins University \cite{he2012toward}, \cite{uneri2010new} has developed a co-manipulation system called the Steady-Hand Eye Robot (SHER) that measures the force applied by the user to filter hand tremor. Gonenc et al. \cite{gonenc2014motorized} integrated a force-sensing motorized micro-forceps with the hand-held device Micron tool to reduce hand tremor. Ida et al. \cite{ida2012microsurgical} designed a teleoperated system to enable microcannulation experiments on an \textit{ex vivo} porcine eye. To detect scleral force, He et al. \cite{he2014multi} developed a multi-functional force-sensing instrument that measured both the scleral and tip forces. A recurrent neural network was further designed and trained to predict future safety status \cite{he2019enabling}. %Gijbels et al. \cite{} realzied hand tremor filtering with the aid of a non-backdrivable device, and achieved the world-first in-human robot-assisted retinal vein cannulation.

\subsection{Depth Estimation}
Research on depth estimation in image guided retinal surgery relies on the feedback of features generated by the light source. Zhou et al. \cite{zhou2020spotlight} have developed a method to measure the distance between the instrument and surface based on projection patterns of the spotlight in the microscope camera. Koyama et al. \cite{koyama2022autonomous} proposed a shadow-based autonomous positioning using two robotic manipulators. They automated the motion of the light guide to ensure that the shadow of the instrument was always inside the microscopic view. Our prior work \cite{kim2020autonomously}, \cite{kim2021towards}, \cite{zhang2021towards} demonstrated that a deep neural network can be trained to predict depth with high accuracy. The work of others has utilized OCT \cite{ourak2019combined}, \cite{weiss2018fast}, \cite{kang2018demonstration}, \cite{cheon2015accurate}, which can provide significantly higher resolution than a microscope camera, but at a higher cost.

\subsection{Head Drift or Eye Movement}

Head drift or eye movement is an underappreciated problem during intraocular surgery. It may significantly increase the challenge or even prevent successful procedure completion. Severe intraoperative complications may result and diminish surgical outcomes \cite{brogan2018intraoperative}. McCannel et al. \cite{mccannel2012snoring} reported that in 20 out of 230 vitreoretinal surgery cases major head movement occurred. Notably, 18 out of 37 snoring patients moved their heads substantially during the surgery. 

\section{PROBLEM SETUP}

To achieve the autonomous navigation process in Fig. \ref{fig:workflow}, we first define the tool-tip state of the surgical tool attached to the robot end-effector as $x=(\boldsymbol{p},R,\boldsymbol{v},\boldsymbol{\omega})$, where $\boldsymbol{p}\in\mathbb{R}^3$ and $R\in SO(3)$ are the tool-tip position and orientation, $\boldsymbol{v}\in\mathbb{R}^3$ and $\boldsymbol{\omega}\in\mathbb{R}^3$ are the linear and angular velocity relative to robot base frame. The control inputs are defined as $u=(\boldsymbol{u}_v, \boldsymbol{u}_\omega)$, where $\boldsymbol{u}_v\in\mathbb{R}^3$ and $\boldsymbol{u}_\omega\in\mathbb{R}^3$ are translational forces  
and joint torques. The full state $x$ can be determined by robot joint encoders and forward kinematics.

Second, we use a convolutional neural network to achieve depth perception. The input of our network combines a RGB top-down image of the current surgical scene and a binary goal position image defined by users. The observed top-down image from a fixed microscope camera at time $t$ is defined as $o(t)\in \mathcal I$. The user is able to specify a 2-D goal position $g_i=(x_i,y_i)\in\mathbb{R}^2$ by clicking directly on the displayed real-time image. We convert this goal position into a binary image and attach as an additional layer along with $o(t)$ to form the input. The output is a 3-D vector defined by $\boldsymbol{d} = [x\ y\ z]^T$, which is a vector predicted by the network from the current tool-tip position to the user-defined goal position. $\|\boldsymbol{d}\|$ is the predicted distance between tool-tip and goal position.

Third, we generate a collision free tool-tip trajectory using chance constraint incorporated DDP. The cost function and constraints over time interval $[t_0,t_f]$ are defined as:
\begin{align}
    &\arg\!\min_{\!\!\!\!\!\!\!\!\!\!\! x(\cdot),u(\cdot)} \int_{t_0}^{t_f} C(x(t), u(t)) dt, \text{\ \ \ \ \ \ subject to:}\label{eq:problem_cost} \\
    &\text{robot dynamics:\ \ \ }\dot x(t) = f(x(t), u(t)), \label{eq:problem_dynamics}\\
    &\text{scleral constraint:\ } (I-\boldsymbol{v}_o(t) \boldsymbol{v}_o(t)^T)(\boldsymbol{p}_s - \boldsymbol{p}(t)) = 0, \label{eq:problem_sclera}\\
    &\text{terminal constraint:\ } x(t_f) \in X_f, \label{eq:problem_terminal}\\
    &\text{collision chance constraint:}\nonumber\\
    &\mathcal Pr\left( \|\boldsymbol{p}(t)-\boldsymbol{p}_c\| \leq l(t) \right) \geq \alpha, \text{ for } 0 \leq \alpha \leq 1,  \label{eq:problem_collision}
\end{align}
where $\boldsymbol{v}_o(t)$ is a unit vector representing the current orientation of surgical tool. $\boldsymbol{p}_s\in\mathbb{R}^3$ is the scleral incision point. The translational motion of the instrument at this point will cause a net force applied to the sclera, which could damage the sclera. The scleral constraint closes the gap between the current and desired orientation of the tool shaft (Fig. \ref{fig:FBG_setup_scleral_incision_intact_eye}b), and guarantees the range of movement within a small threshold. $\boldsymbol{p}_c\in\mathbb{R}^3$ is the estimated ellipsoid center, $\alpha\in\mathbb{R}$ is the level of confidence, and $l(t)\in\mathbb{R}$ is a length parameter defined as:
\begin{gather}
    l(t) = \|\boldsymbol{p}_c - \boldsymbol{p}_{proj}(t)\|,\label{length}
\end{gather}
where $\boldsymbol{p}_{proj}(t)$ is the projection point that projects the corresponding sample point $\boldsymbol{p}(t)$ onto the estimated ellipsoid surface along the direction from $\boldsymbol{p}_c$ pointed to the sample point. For example, if the estimated geometry is a sphere instead of an ellipsoid, $l(t)$ is equal to the radius.

The stop position of the tool-tip should be immediately above the goal position and yet, without contact with the retinal surface. We summarize the process (Fig. \ref{fig:workflow}) as follows: 1) Define a 2-D goal position through mouse clicking. 2) Collect over 200 sample points around the goal position based on predictions of the network. 3) Estimate the local retinal geometry as though it's an ellipsoid using the weighted least squares method. 4) Use DDP-based MPC to navigate the surgical tool to the goal position autonomously and safely.
% \begin{enumerate}
%   \item Define a 2-D goal position through mouse clicking.
%   \item Collect over 200 sample points around the goal position based on predictions of the network.
%   \item Estimate the local retinal geometry as though it's an ellipsoid using the weighted least squares method.
%   \item Use DDP-based MPC to navigate the surgical tool to the goal position autonomously and safely.
% \end{enumerate}

\section{TECHNICAL APPROACH}

\subsection{Experimental Setup}

\begin{figure}[thpb]
  \centering
  \includegraphics[scale=0.072]{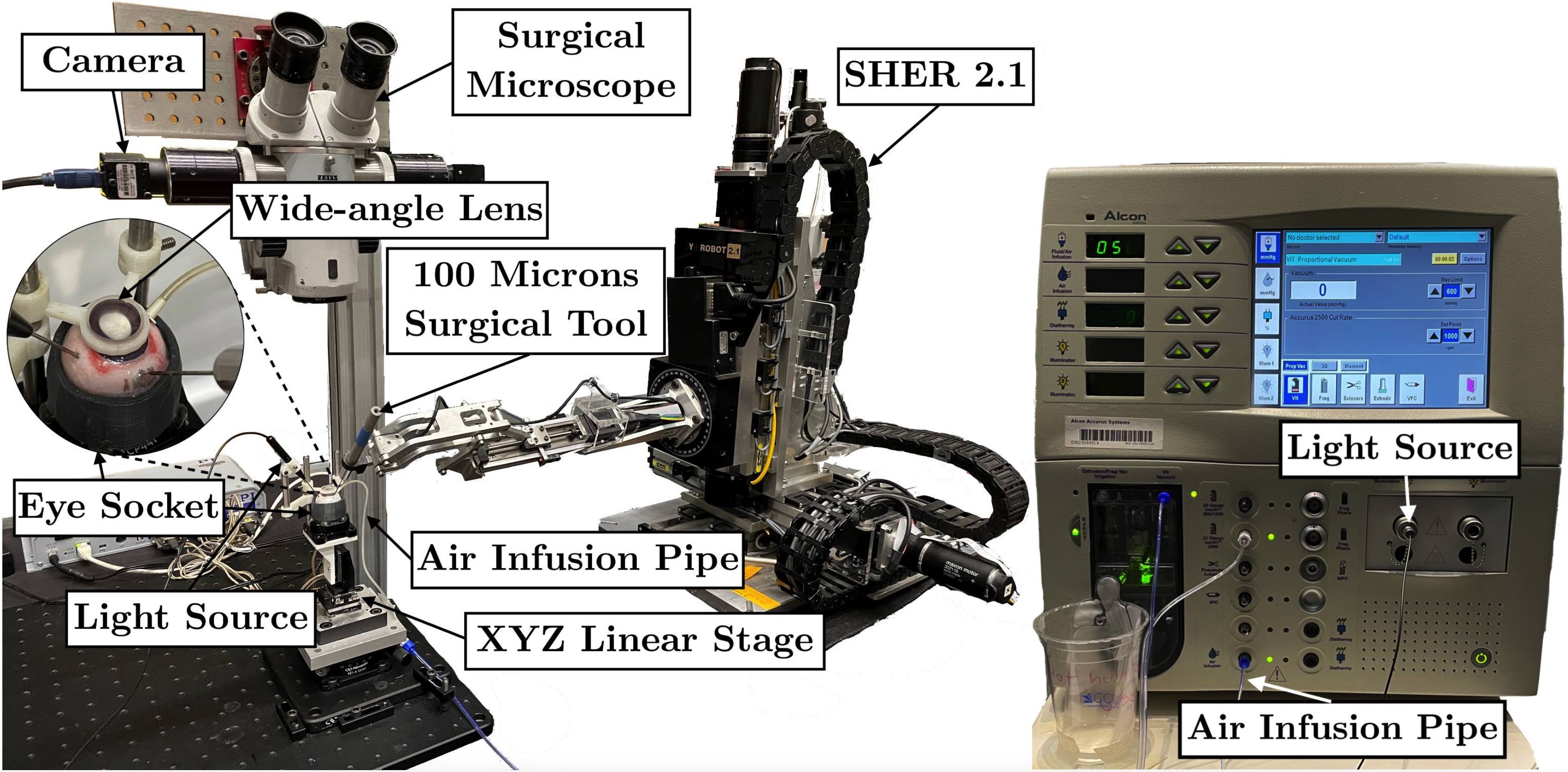}
  \vspace{-5pt}
  \caption{Intact eye experimental setup. The vitrectomy machine provides the light source and air infusion to maintain the intraocular pressure.}
  \label{fig:setup}
  \vspace{-18pt}
\end{figure}

Both open-sky and intact porcine eye experimental setups contain a surgical robot manipulator (SHER 2.0 for open-sky eye and SHER 2.1 for intact eye experiments) with a surgical needle attached to its end-effector, a Q-522 Q-motion XYZ linear stage from Physik Instrumente (PI), an E-873.3QTU linear stage controller (PI), and a microscope camera to obtain the current image. The light source for the open-sky porcine eye experiments is on the right, while we use a brighter light source for the intact porcine eye experiments on the left - to enhance visual feedback inside the eye. An Accurus 800CS vitrectomy machine from Alcon Surgical is used to remove the lens and vitreous humor. The vitrectomy infusion maintains the intraocular pressure for intact eyes, avoiding eye collapse during experiments. A 3D-printed eye socket is used to secure and stabilize the eye. A wide-angle lens is used to provide a wide field of view and to simulate the distortion caused by the human lens, Fig. \ref{fig:setup}. The balanced saline solution is used to couple the gap between the lens and cornea. The outer diameter of the tool-tip is 0.1 mm.

\subsection{Eye Preparation}
The most time-consuming part before data collection is preparation of suitable intact eyes, that allows a clear top-down camera view. Retinal detachment and corneal opacity are two of the main obstacles in carrying out the experiments. While the intact eye allows immediate capture of a clear view through the lens and vitreous humor directly, the view is lost relatively quickly. As it requires one hour to collect 150 trajectories for one eye, we remove the lens and leave most vitreous humor inside the eye. Maintaining corneal clarity remains a challenge that is mitigated by keeping the cornea wet and utilizing the freshest eyes obtainable.

\subsection{Data Collection}
1800 trajectories are collected from 12 different porcine eyes (150 for each) for open-sky eye experiments, and 2100 trajectories are collected from 14 different porcine eyes for the intact eye experiments. Since the geometry of fresh cadaveric porcine eye varies from eye-to-eye, we collect the training data manually. To collect one trajectory for each experiment, we start the surgical tool from a random position and navigate it to a random goal position in a straight line. Both current surgical scene images and tool-tip positions are recorded at a frequency of 15 fps. We manually label the tool-tip stop position of the last frame from each trajectory, and generate the binary goal image based on this. The direction and distance vector label are calculated by subtracting the last frame tool-tip position from other frames.

\subsection{Network Training}

We build our network based on ResNet-18 model \cite{he2016deep}. 75\% of collected trajectories are used for training and the rest are used as a validation dataset. We discretize continuous XYZ coordinates into bins to better label distance vectors. Each bin represents 0.02 mm. In this work, we divide XYZ axes into 854, 887, and 360 bins for both experiments. The cross-entropy loss is used as the loss function.

\subsection{Regularized Geometry Fitting}
We use weighted least squares ellipsoid estimation for local retinal geometry fitting. The ellipsoid fitting is a continuation of works by Li \cite{li2004least} and Reza \cite{reza2017least}. We add a regularized term to the generalized ellipsoid equation, which fits better to our model with some prior information:
\begin{align}
    \ \ a&x^2+by^2+cz^2+2fyz+2gxz \nonumber\\
    &+2hxy+2px+2qy+2rz+d \nonumber\\
    +\lambda[(x-&x_0)^2+(y-y_0)^2+(z-z_0)^2-r_0^2]=0, \label{eq:ellipsoid_general} \\
    &\ \ \ \text{subject to}\ \ \ \ kJ-I^2=1, \nonumber
\end{align}
where $I = a + b + c, J = ab + bc + ac - f^2 - g^2 - h^2$, $\lambda$ is the regularization gain, $x_0$, $y_0$, $z_0$, and $r_0$ are the ellipsoid center position and radius based on prior information. Let:
\begin{gather}
    X_i = (x_i^2,y_i^2,z_i^2,2y_iz_i,2x_iz_i,2x_iy_i,2x_i,2y_i,2z_i,1)^T, \nonumber\\
    \boldsymbol{v} = (a+\lambda,b+\lambda,c+\lambda,f,g,h,p-\lambda x_0,q-\lambda y_0, \nonumber \\ ,r-\lambda z_0,d+\lambda[x_0^2+y_0^2+z_0^2-r_0^2])^T, \nonumber
\end{gather}
Equation \eqref{eq:ellipsoid_general} can be written with weight matrix $W$ as:
\begin{gather}
    min\ \boldsymbol{v}^TDWD^T\boldsymbol{v} \ \ \ \ \text{subject to}\ \ \ \ \ \boldsymbol{v}^TC\boldsymbol{v}=1, \label{eq:ellipsoid_lagrange} \\
    W =\begin{bmatrix}
    w_1 & 0 & \cdots & 0\\
    0 & w_2 & \cdots & 0\\
    \vdots & \vdots & \ddots & \vdots\\
    0 & 0 & \cdots & w_n\\
    \end{bmatrix} = \begin{bmatrix}
    \frac{1}{\sigma_1^2} & 0 & \cdots & 0\\
    0 & \frac{1}{\sigma_2^2} & \cdots & 0\\
    \vdots & \vdots & \ddots & \vdots\\
    0 & 0 & \cdots & \frac{1}{\sigma_n^2}\\
    \end{bmatrix},\nonumber
\end{gather}
where $D = (X_1,X_2,\cdots,X_n)\in \mathbb{R}^{10\times n}$, and $\sigma_n^2$ is the distance variance calculated from the validation dataset.

Using the Lagrange multiplier method, $\boldsymbol{v}$ is solved as an eigenvalue and eigenvector problem. We can rewrite \eqref{eq:ellipsoid_general} as:
\begin{gather}
    (\boldsymbol{x} - \boldsymbol{p}_c)^TLL^T(\boldsymbol{x} - \boldsymbol{p}_c)=1,
\end{gather}
where $\boldsymbol{p}_c=[x_0\ y_0\ z_0]^T$ is the estimated center of ellipsoid, and $L$ is a lower triangular matrix.

Then \eqref{eq:problem_collision} can be expressed as:
\begin{gather}
    \mathcal Pr\left( (\boldsymbol{p}(t)-\boldsymbol{p}_c)^TLL^T(\boldsymbol{p}(t)-\boldsymbol{p}_c) \leq 1 \right) \geq \alpha,
    \label{eq:ellipsoid_chance}
\end{gather}

\subsection{Chance Constraint Incorporated DDP and MPC}
It is difficult to incorporate the chance constraint into the cost function directly. However, Blackmore et al. \cite{blackmore2011chance} has proved that a linear chance constraint is equal to a deterministic constraint on the mean of x denoted as:
\begin{gather}
    \mathcal Pr\left( \boldsymbol{a}^T\boldsymbol{x} \leq b\right) \leq \delta\  \Longleftrightarrow \ \boldsymbol{a}^T\hat{\boldsymbol{x}} - b \geq c, \label{eq:chance_translate}
\end{gather}
where $\boldsymbol{x} \sim \mathcal{N} (\hat{\boldsymbol{x}}, \Sigma)$ is a multivariate random variable with fixed covariance, $\delta$ is the level of confidence, $c = erf^{-1}(1- 2\delta)\sqrt{2\boldsymbol{a}^T\Sigma \boldsymbol{a}}$, and $erf(x)=\frac{2}{\sqrt{\pi}}\int_{0}^{x} e^{-t^2}dt$ is the standard error function. 

Instead of approximating the nonlinear constraint at the center only in our previous work \cite{zhang2021towards}, we linearize \eqref{eq:ellipsoid_chance} at both $\hat{\boldsymbol{p}_c}$ and $\hat{L}$. $\hat{\boldsymbol{p}_c}$ and $\hat{L}$ can be derived by running geometry estimation several times. Then \eqref{eq:ellipsoid_chance} can be written as:
\begin{gather}
    Pr\left( (\boldsymbol{p}(t) - \hat{\boldsymbol{p}_c})^T\hat{L}\hat{L}^T(\boldsymbol{p}(t) - \hat{\boldsymbol{p}_c}) \right. + 2\sum_{i=1}^{3} \sum_{j=1}^{3} m_{ij} -  \nonumber \\
    \left.2(\boldsymbol{p}(t) - \hat{\boldsymbol{p}_c})^T\hat{L}\hat{L}^T(\boldsymbol{p}_c - \hat{\boldsymbol{p}_c}) \leq 1 \right) \geq \alpha,
\end{gather}
where $m_{ij}$ is the element of matrix: 
\begin{gather}
    \left[(\boldsymbol{p}(t) - \hat{\boldsymbol{p}_c})(\boldsymbol{p}(t) - \hat{\boldsymbol{p}_c})^T\hat{L}\right]\odot(L - \hat{L})\in \mathbb{R}^{3\times 3}, \nonumber \\
    \text{let }\ \ \sum_{i=1}^{3} \sum_{j=1}^{3} m_{ij} = K(S - \hat{S}), \nonumber
\end{gather}
where $K\in \mathbb{R}^{1\times 6}$ derives from $\left[(\boldsymbol{p}(t) - \hat{\boldsymbol{p}_c})(\boldsymbol{p}(t) - \hat{\boldsymbol{p}_c})^T\hat{L}\right]$, $S\in \mathbb{R}^{6\times 1}$ derives from $L$, then:
\begin{gather}
    Pr\left(2M\boldsymbol{p}_c + 2KS \leq 1 + b \right) \geq \alpha, 
    \label{eq:lineraized_chance}
\end{gather}
where $M = -(\boldsymbol{p}(t) - \hat{\boldsymbol{p}_c})^T\hat{L}\hat{L}^T$, and $b = -2(\boldsymbol{p}(t) - \hat{\boldsymbol{p}_c})^T\hat{L}\hat{L}^T\hat{\boldsymbol{p}_c} + 2K\hat{S} - (\boldsymbol{p}(t) - \hat{\boldsymbol{p}_c})^T\hat{L}\hat{L}^T(\boldsymbol{p}(t) - \hat{\boldsymbol{p}_c})$. Let $\boldsymbol{a}^T = [2M \ \ 2K]$ and $\boldsymbol{y} = [\boldsymbol{p}_c \ \ S]^T$, then \eqref{eq:lineraized_chance} is euqal to:
\begin{gather}
    % Pr\left( \boldsymbol{a}^T\boldsymbol{y} \leq 1 + b \right) =
    \frac{1}{2} + \frac{1}{2}erf\left( \frac{1 + b - \boldsymbol{a}^T\hat{\boldsymbol{y}}}{\sqrt{2\boldsymbol{a}^T\Sigma \boldsymbol{a}}} \right) \geq \alpha, \nonumber\\
    erf^{-1}(2\alpha - 1)\sqrt{2\boldsymbol{a}^T\Sigma \boldsymbol{a}} + \boldsymbol{a}^T\hat{\boldsymbol{y}} - (1 + b) \leq 0, \nonumber
\end{gather}
where $\Sigma \in \mathbb{R}^{9\times 9}$ is a combination of $\Sigma_{p_c}$ and $\Sigma_{L}$. If $\alpha = 0.99$, then $erf^{-1}(2\alpha - 1) \approx 1.64497636$.

MPC is then implemented by generating trajectories using chance constraint incorporated DDP along the actual navigation track. The time horizon is set to be 5s. Based on each trajectory generated by DDP, we navigate towards the predicted goal for a small step. Then we predict the goal position once again with the current top-down image and use this newly predicted goal to generate a new trajectory using DDP. We continue navigating the surgical tool for each small step and repeating the whole process until the predicted distance is less than 0.1 mm.

\subsection{Optical Flow}
We simulate the head drift by moving the XYZ linear stage randomly along the XY axes. Since our goal position is selected via mouse click on the monitor, the clicked goal on the screen will not move along with the linear stage movement. We must therefore track the goal position movement based on the top-down images. Optical flow is a feasible way to solve this. We first capture a $300 \times 160$ pixel image area before and after the linear stage movement. Then we use the Shi-Tomasi Corner Detector \cite{shi1994good} to detect relevant features in both images. Then we use the Lucas-Kanade method \cite{lucas1981iterative} to derive the resulting motion between the two images. The goal position on the screen and the RCM point are updated based on these calculations.

\subsection{Scleral Force Measurement}

We use a Fiber Brag Grating (FBG) sensors integrated surgical tool (Fig. \ref{fig:FBG_setup_scleral_incision_intact_eye}a) to measure the scleral force during the navigation process. Previous work \cite{he2014multi} has related the raw wavelength data of FBGs to the scleral force along XY axes. The measurement starts from the initial position and ends by retracting the surgical tool to the initial position again. Before each trial, we rebias the force reading at the initial position. After that, we do some preliminary navigation tests to make sure when we retract the tool, the force reading is returned to the original level (i.e., the force reading should close to 0 mN before and after each test). In this work, we want to prove that by using scleral constraint incorporated control, we could keep the scleral forces at the scleral incision point within a safe limit. We set this safe limit as 115 mN based on mean forces using manual manipulators in \textit{in vivo} rabbit experiments performed by two retina specialists. \cite{urias2020robotic}.

\section{EXPERIMENTS}
% Since our goal was to improve depth perception, reduce scleral forces, and achieve autonomous navigation during retinal microsurgery, the evaluation metrics for the experiments focused on these aspects. For both open-sky and intact porcine eye experiments, we evaluated the mean XY axes errors along with mean scleral errors.
\subsection{Open-sky Porcine Eye Experiments}

We used the open-sky porcine eye to evaluate the performance of our network with\begin{figure}[thpb]
  \vspace{-5pt}
  \centering
  \includegraphics[scale=0.091]{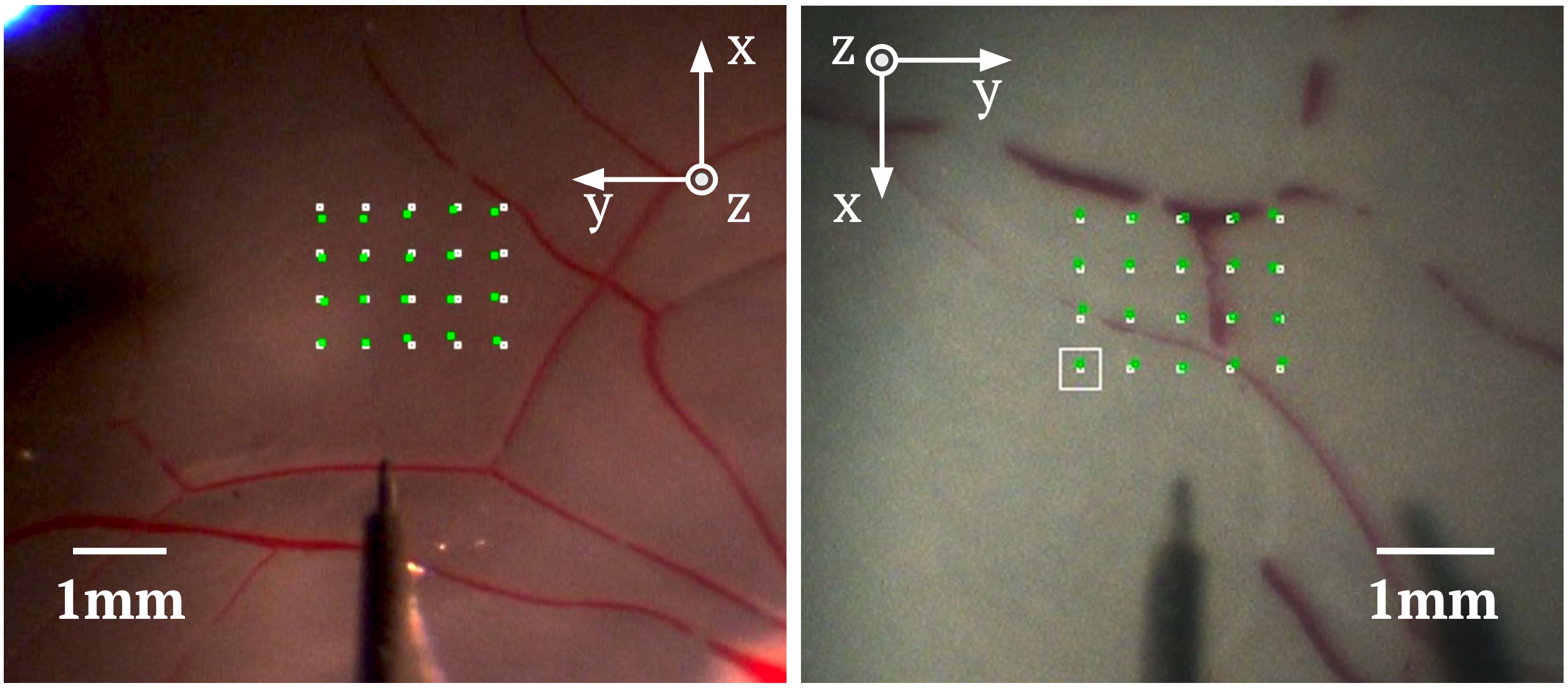}
  \vspace{-5pt}
  \caption{Experimental results for the open-sky eye (left) and the intact eye (right). The white points are pre-defined goal positions and the green points are actual landing positions. The white square is the current goal position. The directions of XY axes are different due to the utilization of different SHERs for the two experiments.}
  \label{fig:20_results}
\end{figure}
\begin{figure*}[thpb]
  \centering
  \includegraphics[scale=0.176]{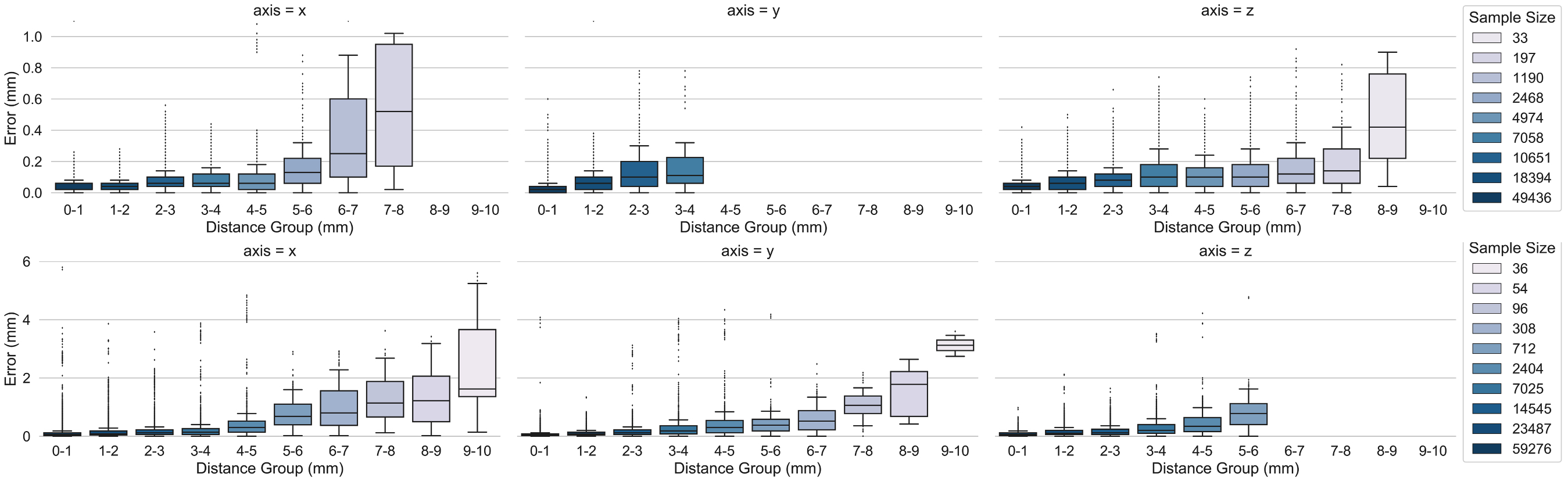}
  \vspace{-10pt}
  \caption{Relationship between distance groups and validation errors by three axes. (top) Open-sky eye (450 trajectories). (bottom) Intact eye (525 trajectories).}
  \label{fig:box_plot}
  \vspace{-5pt}
\end{figure*}
\begin{table*}
% \vspace{-3pt}
\caption{Experimental results.}
\vspace{-12pt}
\label{table:exp_results}
\begin{center}
\begin{tabular}{c|c|c|c|c|c}
\hline
\multicolumn{1}{p{5.6cm}|}{\centering evaluation items \\ (units)}
& \multicolumn{1}{p{1.9cm}|}{\centering \# of tested goal positions}
& \multicolumn{1}{p{1.7cm}|}{\centering mean X error\\(mm)}
& \multicolumn{1}{p{1.7cm}|}{\centering mean Y error\\(mm)}
% & \multicolumn{1}{p{1cm}|}{\centering z error \\ (mm)}
& \multicolumn{1}{p{2.3cm}|}{\centering mean time duration\\(s)}
& \multicolumn{1}{p{1.8cm}}{\centering \# of trajectories \\ hit the retina} \\
% evaluation items & x error & y error & z error & scleral error & depth score\\
\hline
open-sky-eye1-without-chance-constraint & 20 & 0.0513 & 0.0463 & N/A & 10\\
\hline
open-sky-eye1-chance-constraint (Fig. \ref{fig:20_results}) & 20 & 0.0525 & 0.0463 & N/A & 7\\
\hline
open-sky-eye2-without-chance-constriant & 20 & 0.0413 & 0.0475 & N/A & 11\\
\hline
open-sky-eye2-chance-constraint & 20 & 0.0600 & 0.0400 & N/A & 5\\
\hline
intact-eye3-without-chance-constraint & 20 & 0.0286 & 0.0429 & 36.194 & 10\\
\hline
intact-eye3-MPC-100$\mu$m-stop-condition (Fig. \ref{fig:20_results}) & 20 & 0.0253 & 0.0374 & 10.569 & 0\\
\hline
intact-eye3-MPC-300$\mu$m-stop-condition & 20 & 0.0220 & 0.0220 & 7.208 & 0\\
\hline
intact-eye3-head-drift-100$\mu$m-stop-condition & 20 & 0.0352 & 0.0385 & 15.021 & 0\\
\hline
intact-eye3-head-drift-300$\mu$m-stop-condition & 20 & 0.0451 & 0.0440 & 10.809 & 0\\
\hline
\end{tabular}
\end{center}
\vspace{-25pt}
\end{table*}and without a chance constraint. For each trajectory, we navigated the surgical tool directly towards the goal until the tool-tip was 2mm away from the goal position. Then we followed the DDP-generated trajectory to reach the goal. Two eyes were tested. For each eye, we navigated the surgical tool to 20 pre-defined positions autonomously. The 20 positions were selected from a $60 \times 80$ pixel area in the camera view, Fig.  \ref{fig:20_results}. The actual size of the test area selected was $1.5\times2.0$ mm. The determination of depth perception performance was evaluated based on the number of trajectories that hit the retinal surface. This was done by assessing light reflection patterns around the contact area via the recorded video.

\subsection{Intact Porcine Eye Experiments}

In the intact porcine eye experiments, we first navigated to 20 pre-defined goal positions ($60 \times 80$ pixel) without using chance constraint. We used this result as a baseline. Then we navigated to the same 20 pre-defined goal positions using MPC. The actual size of the test area was $1.32\times1.76$ mm due to the different magnifications of microscope. The tool was considered to reach the goal if the norm of the network prediction was smaller than 0.1 mm. To make a comparison, we added another stop condition and repeated the experiment. The tool will be stopped as well if the norm of the network prediction was smaller than 0.3 mm and the norm of tool tip error along XY axes was smaller than 2.3 pixels. Finally, we tested the performance of our system in the presence of modeled head drift using the two stop conditions mentioned above. We controlled the XYZ linear stage to randomly move along XY axes for a small distance during each navigation process. The movement range for each axis was from [-0.25 mm, 0.25 mm]. Head drift was only simulated within this small range. Otherwise, we lost the working area view from the fixed camera. Also, the scleral forces applied to the scleral wall were potentially too large to prevent surgical tool bending. When there was residual vitreous humor in the eye, we were unable to evaluate the performance of depth perception using the reflection of light. In this instance we assessed tool-to-retina contact by retinal surface deformation at the user-defined goal position.

\subsection{Scleral Force Measurement}
We navigated the FBG sensors integrated tool to 10 pre-defined positions to measure the scleral force during the MPC navigation. To evaluate the head drift motion, we navigated to a fixed position 10 times. For each time, we randomly moved the linear stage for a small distance along XY axes. We increased the movement range for each axis to [-0.5 mm, 0.5 mm] to make the change of force more visible. All goal positions were manually defined above the retinal surface.

\section{RESULTS}
For the best network model of open-sky eye experiments, the mean training errors for XYZ axes were 0.028, 0.016, and 0.042 mm and the mean validation errors for XYZ axes were 0.058, 0.044, and 0.087 mm respectively. For the best network model of intact eye experiments, the mean training errors for XYZ axes were 0.049, 0.041, and 0.054 mm and the mean validation errors for XYZ axes were 0.152, 0.102, and 0.157 mm respectively. The relationship between distance groups and prediction errors was plotted in Fig. \ref{fig:box_plot}. A trend was evident indicating that when the tool was closer to the retina, the prediction errors were smaller.
\begin{figure}[thpb]
  \centering
  \includegraphics[scale=0.085]{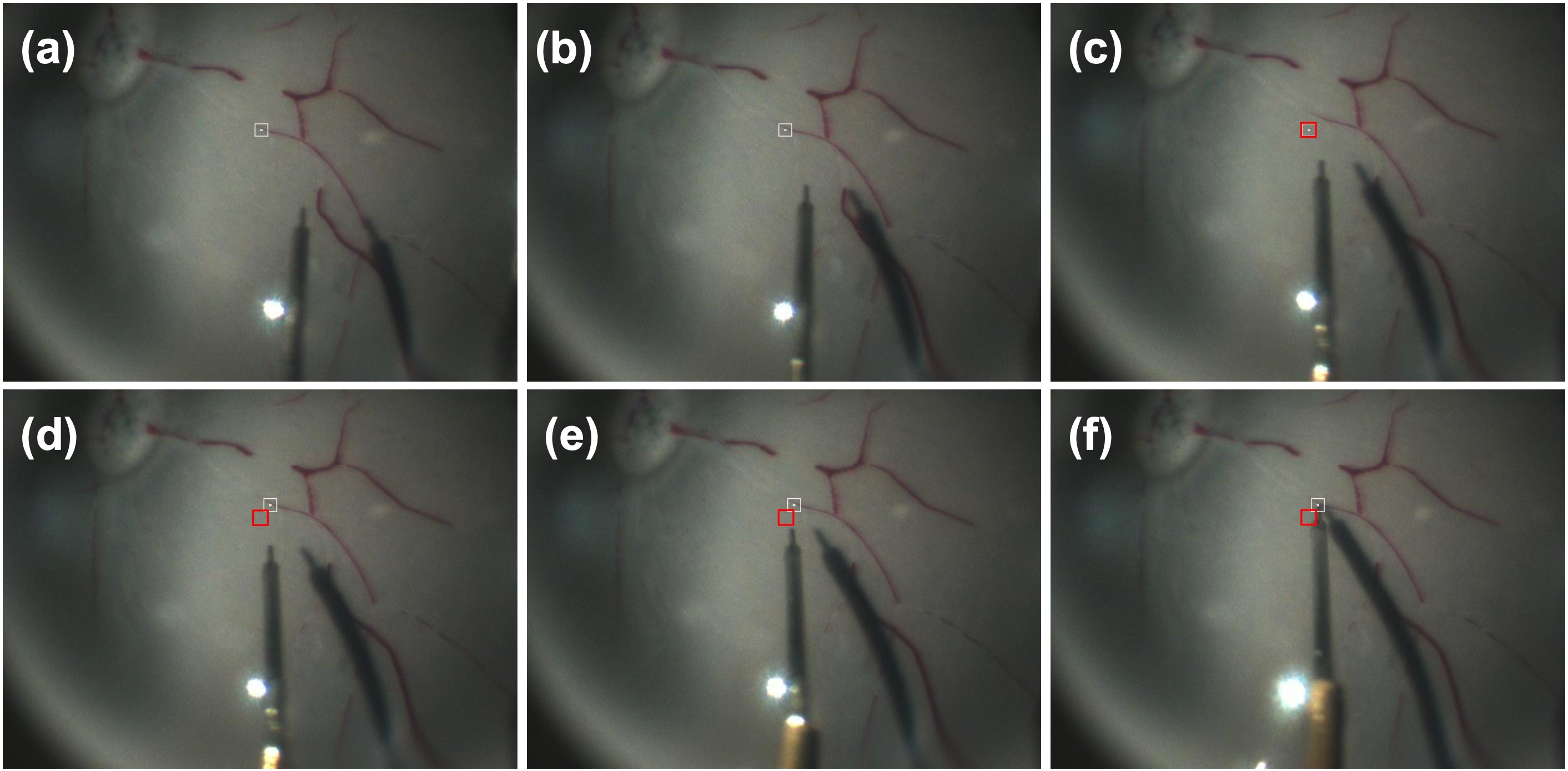}
  \vspace{-5pt}
  \caption{(a)-(f) Head drift simulation results. The head drift happens between (b) and (c). The white square in (a)-(c) is the original user-defined goal, and the white square in (d)-(f) is the updated goal. The red square represents the original goal after the head drift. In (c) the overlap occurs when the head drift happens, but the goal has not been updated.}
  \label{fig:head_drift}
  \vspace{-15pt}
\end{figure}
The validation errors of the intact eye were larger than open-sky eye's due to increased blur in these top-down training images. This was not avoidable as in the time required for data collection degradation of the view through the intact cornea was progressive. Despite corneal opacity the prediction accuracy of the last few millimeters remained relevant. Fig. \ref{fig:box_plot} showed that it was in an acceptable range of approximately 0.05 mm. The mean X and Y errors in Table \ref{table:exp_results} were mean errors between pre-defined goal positions and actual tool-tip stop positions in the top-down image, Fig. \ref{fig:20_results}.

\subsection{Open-sky Porcine Eye Experiments}
The last column of Table \ref{table:exp_results} demonstrated that by using the chance constraint, the result was a more reliable mean to determine depth. Although tool-to-retina contact had occurred during these experiments, it was not sufficient to cause retinal tissue deformation or damage. It was noted that the chance constraint method rendered the autonomous navigation process safer, but this did come at the cost of a small reduction in accuracy along the XY axes.
% The final landing positions of two different porcine eyes are plotted in Fig.\ref{fig:sample_results}.

\subsection{Intact Porcine Eye Experiments}
The results of intact eye experiments showed that using the MPC can almost eliminate the collision risk between the tool-tip and the retinal surface. This was because the MPC kept predicting the goal position until the tool-tip was very close to the goal. This differed from the open-sky eye experiments where the last prediction occurred at 2 mm from the goal position. This fact contributed to smaller mean XY axes errors. Also, the mean time duration was reduced significantly from 36.194s to 7.208s by using MPC. Fig. \ref{fig:head_drift} showed one trial of the head drift simulation. The last two rows in Table \ref{table:exp_results} showed that the surgical tool was able to track the updated goal positions following head drift for all 40 trials. The mean navigation time of head drift was longer than MPC.
\begin{figure}[thpb]
  \centering
  \includegraphics[scale=0.095]{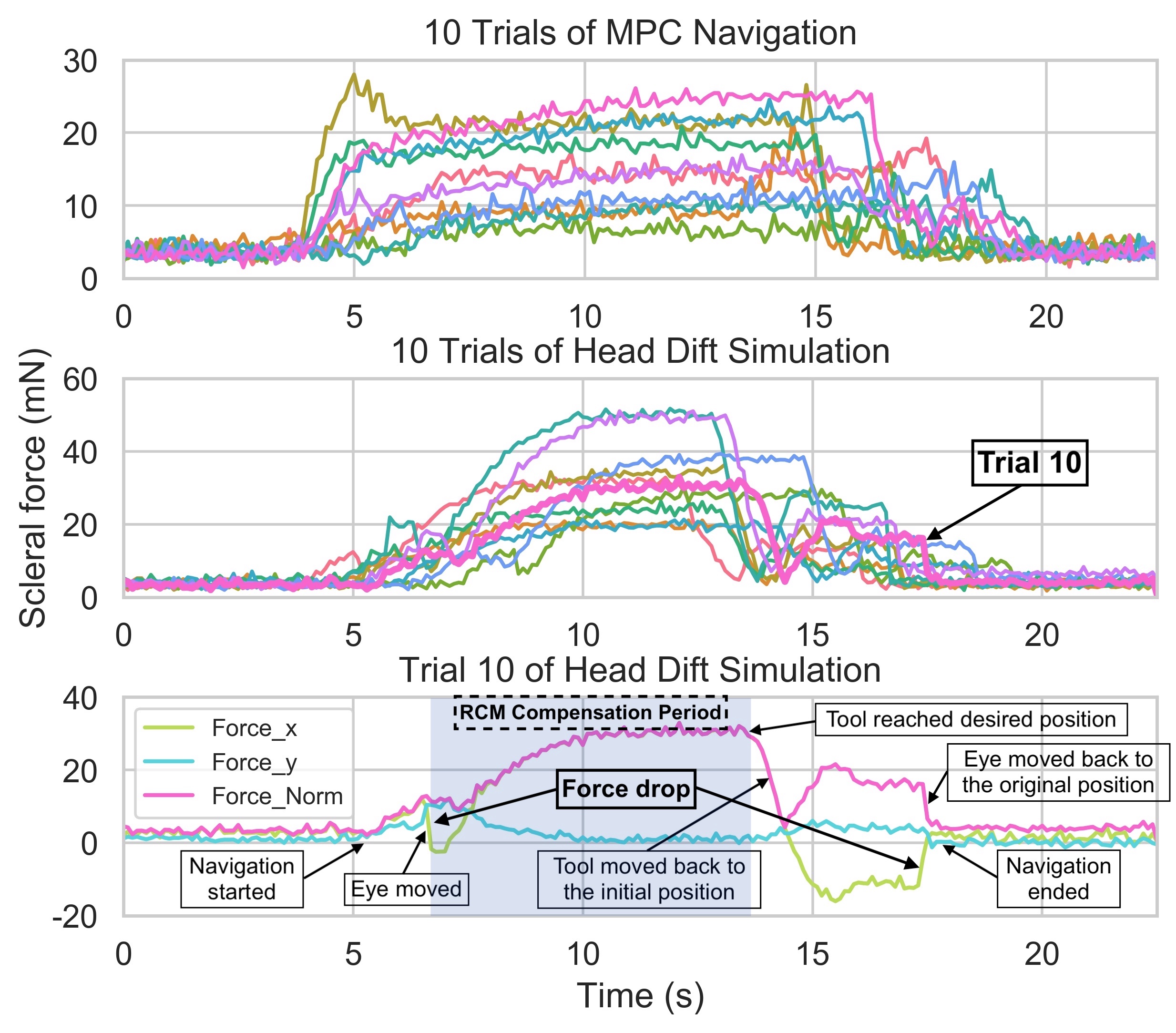}
  \vspace{-12pt}
  \caption{Results of scleral force measurement. (Top) Scleral force of 10 trials of MPC navigation. (Middle) Scleral force of 10 trials of head drift simulation. (Bottom) Trial 10 0f head drift simulation.}
  \label{fig:scleral_force}
  \vspace{-15pt}
\end{figure}
This may result from the time spent to relocate the updated goal, and changes in the camera view and the relative shadow position. Experiments with 0.1 mm stop condition had longer mean time duration than the 0.3 mm stop condition, as for some trials the tool paused for the last few seconds waiting for the network prediction to meet the stop condition.

\subsection{Scleral Force Measurement}
Fig. \ref{fig:scleral_force} showed the experimental results from the scleral force measurements. As aforementioned, we set the safe limit of the scleral force to be 115 mN based on the work of Urias \cite{urias2020robotic}.  The results showed that no measured forces exceeded 30 mN for MPC navigation or 60 mN during head drift motion. The mean scleral forces were 11.97 mN and 21.75 mN respectively. Take trial 10 of head drift simulation as an example, the scleral force changed drastically along one direction when the eye moved. Since we updated the RCM point based on the movement of eye, the scleral force would drop a little bit afterwards. This RCM compensation lasted until we reached the goal (The light blue area in Fig. \ref{fig:scleral_force}). As the tool moved towards the goal, it also rotated to reach the desired orientation. This further increased the scleral force until the goal was reached. Subsequently, we retracted the tool to the initial position using the original RCM point. Finally, we moved the eye back to the original position, explaining the large change of scleral force at the end of the trial. The mean scleral force was calculated from navigation start to the moment that the tool reached the goal.

\section{CONCLUSIONS}

In this work, we test our improved robotic system using both open-sky and intact \textit{ex vivo} porcine eyes. The results demonstrate smaller than 0.06 mm mean XY axes errors during experiments involving biological tissues. Moreover, by using chance constraint incorporated MPC, a safer, faster, and more accurate navigation process is obtained. We also demonstrate that our system is sufficiently robust to track the updated goal accurately during simulated head drift experiments. However, we only simulated head drift within a limited range. More work is required to refine performance especially during head movement. Future work will focus on experiments that use intact porcine eyes for the surgically relevant task of subretianl injection. 


%%%%%%%%%%%%%%%%%%%%%%%%%%%%%%%%%%%%%%%%%%%%%%%%%%%%%%%%%%%%%%%%%%%%%%%%%%%%%%%%

%%%%%%%%%%%%%%%%%%%%%%%%%%%%%%%%%%%%%%%%%%%%%%%%%%%%%%%%%%%%%%%%%%%%%%%%%%%%%%%%


% N/A

\section*{ACKNOWLEDGMENT}

This work was supported by U.S. National Institutes of Health under the grants number 2R01EB023943-04A1, 1R01EB025883-01A1, and partially by JHU internal funds.

%%%%%%%%%%%%%%%%%%%%%%%%%%%%%%%%%%%%%%%%%%%%%%%%%%%%%%%%%%%%%%%%%%%%%%%%%%%%%%%%
\bibliographystyle{IEEEtran}
\bibliography{references}  % .bib

\end{document}